\def\year{2020}\relax
\documentclass[letterpaper]{article} 
\usepackage{aaai20}  
\usepackage{times}  
\usepackage{helvet}  
\usepackage{courier}  
\usepackage{url}  
\usepackage{graphicx}  
\usepackage{xcolor}
\usepackage{fancyvrb}
\usepackage{csquotes}
\usepackage{titlesec}
\begin{document}
%
\title{Common-Knowledge Concept Recognition for SEVA} 

\author{Jitin Krishnan\textsuperscript{1},
Patrick Coronado\textsuperscript{2},
Hemant Purohit\textsuperscript{3},
Huzefa Rangwala\textsuperscript{4}\\
\textsuperscript{1,4}{Department of Computer Science, George Mason University}\\
\textsuperscript{2}{Instrument Development Center, NASA Goddard Space Flight Center}\\
\textsuperscript{3}{Information Sciences \& Technology Department, George Mason University }\\
jkrishn2@gmu.edu,
patrick.l.coronado@nasa.gov,
hpurohit@gmu.edu,
rangwala@gmu.edu}

\maketitle
\begin{abstract}
We build a common-knowledge concept recognition system for a Systems Engineer's Virtual Assistant (SEVA) which can be used for downstream tasks such as relation extraction, knowledge graph construction, and question-answering. The problem is formulated as a token classification task similar to named entity extraction. With the help of a domain expert and text processing methods, we construct a dataset annotated at the word-level by carefully defining a labelling scheme to train a sequence model to recognize systems engineering concepts. We use a pre-trained language model and fine-tune it with the labeled dataset of concepts. In addition, we also create some essential datasets for information such as abbreviations and definitions from the systems engineering domain. Finally, we construct a simple knowledge graph using these extracted concepts along with some hyponym relations. 
\end{abstract}

\subsubsection*{Keywords:} Natural Language Processing, Named Entity Recognition, Concept Recognition, Relation Extraction, Systems Engineering.

\section{INTRODUCTION}

The Systems Engineer's Virtual Assistant (SEVA) \cite{seva} was introduced with the goal to assist systems engineers (SE) in their problem-solving abilities by keeping track of large amounts of information of a NASA-specific project and using the information to answer queries from the user. In this work, we address a system element by constructing a common-knowledge concept recognition system for improving the performance of SEVA, using the static knowledge collected from the Systems Engineering Handbook \cite{SEHandbook} that is widely used in projects across the organization as domain-specific commonsense knowledge. At NASA, although there exists knowledge engines and ontologies for the SE domain such as MBSE \cite{mbse1}, IMCE \cite{imce}, and OpenCaesar \cite{opencaesar}, generic commonsense acquisition is rarely discussed; we aim to address this challenge. 

SE commonsense comes from years of experience and learning which involves background knowledge that goes beyond any handbook. Although constructing an assistant like SEVA system is the overarching objective, a key problem to first address is to extract elementary common-knowledge concepts using the SE handbook and domain experts. We use the term \textit{`common-knowledge'} as the \textit{`commonsense'} knowledge of a specific domain. This knowledge can be seen as a pivot that can be used later to collect \textit{`commonsense'} knowledge for the SE domain. We propose a preliminary research study that can pave a path towards a comprehensive commonsense knowledge acquisition for an effective Artificial Intelligence (AI) application for the SE domain. Overall structure of this work is summarized in Figure 1. Implementation with demo and dataset is available at: \begin{small} {\color{blue} \url{https://github.com/jitinkrishnan/NASA-SE}} \end{small}.

\begin{figure}
  \centering
    \includegraphics[width=6.8cm]{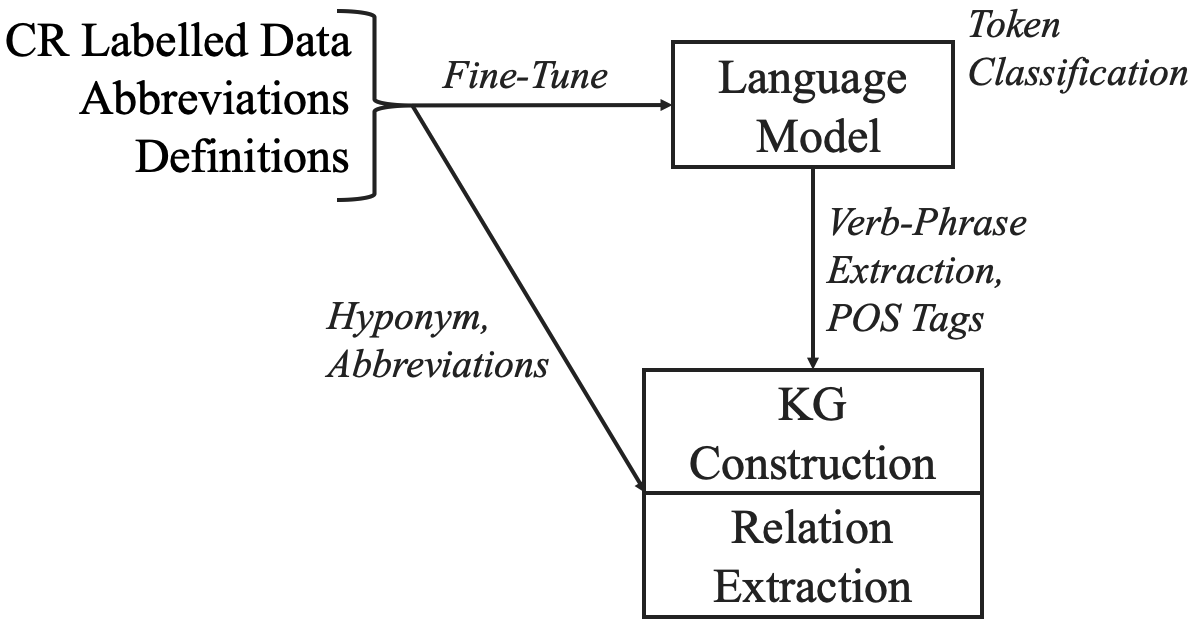}
 \caption{Common-knowledge concept recognition and simple relation extraction}
\end{figure}

\section{BACKGROUND AND MOTIVATION}

Creating commonsense AI still remains an important and challenging task in AI research today. Some of the inspiring works are the CYC project \cite{Panton2006} that tries to serve as a foundational knowledge to all systems with millions of everyday life commonsense assertions,  Mosaic Commonsense Knowledge Graphs and Reasoning \cite{Zellers2018SWAGAL} that addresses aspects like social situations, mental states, and causal relationships, and Aristo System \cite{Aristo} that focuses on basic science knowledge. In NASA's context, systems engineering combines several engineering disciplines requiring extreme coordination and is prone to human errors. This, in combination with the lack of efficient knowledge transfer of generic lessons-learned makes most technology-based missions risk-averse. Thus, a comprehensive commonsense engine can significantly enhance the productivity of any mission by letting the experts focus on what they do best.

Concept Recognition (CR) is a task identical to the traditional Named Entity Recognition (NER) problem. A typical NER task seeks to identify entities like name of a person such as \textit{`Shakespeare'}, a geographical location such as \textit{`London'}, or name of an organisation such as \textit{`NASA'} from unstructured text. A supervised NER dataset consists of the above mentioned entities annotated at the word-token level using labelling schemes such as BIO which provides beginning (B), continuation or inside (I), and outside (O) representation for each word of an entity. \cite{baevski2019cloze} is the current top-performing NER model for CoNLL-2003 shared task \cite{sang2003introduction}. Off-the-shelf named entity extractors do not suffice in the SE common-knowledge scenario because the entities we want to extract are domain-specific concepts such as \textit{`system architecture'} or \textit{`functional requirements'} rather than physical entities such as \textit{`Shakespeare'} or \textit{`London'}. This requires defining new labels and fine-tuning. 

Relation extraction tasks extract semantic relationships from text. These extractors aim to connect named entities such as \textit{`Shakespeare'} and \textit{`England'} using relations such as \textit{`born-in'}. Relations can be as simple as using hand-built patterns or as challenging as using unsupervised methods like Open IE \cite{openie}; with bootstrapping, supervised, and semi-supervised methods in between. \cite{xu2019connecting} and \cite{soares2019matching} are some of the high performing models that extract relations from New York Times Corpus \cite{riedel2010modeling} and TACRED challenges \cite{zhang2017tacred} respectively. Hyponyms represent hierarchical connection between entities of a domain and represent important relationships. For instance, a well-known work by \cite{hearst1992automatic} uses syntactic patterns such as [\textit{Y such as A, B, C}], [\textit{Y including X}], or [\textit{Y, including X}] to extract hyponyms. Our goal is to extract preliminary hyponym relations from the concepts extracted by the CR and to connect the entities through verb phrases. 

\section{CONCEPT RECOGNITION}

SE concepts are less ambiguous as compared to generic natural language text. A word usually means one concept. For example, the word \textit{`system'} usually means the same when referring to a \textit{`complex system'}, \textit{`system structure'}, or \textit{`management system'} in the SE domain. In generic text, the meaning of terms like \textit{`evaluation'}, \textit{`requirement'}, or \textit{`analysis'} may contextually differ. We would like domain specific phrases such as \textit{`system evaluation'}, \textit{`performance requirement'}, or \textit{`system analysis'} to be single entities. Based on the operational and system concepts described in \cite{seva}, we carefully construct a set of concept-labels for the SE handbook which is shown in the next section. 

\subsection{BIO Labelling Scheme}

\begin{enumerate}
    \itemsep0em
    \item \textbf{\textit{abb}}: represents abbreviations such as \textit{TRL} representing \textit{Technology Readiness Level}. 
    \item \textbf{\textit{grp}}: represents a group of people or an individual such as \textit{Electrical Engineers},  \textit{Systems Engineers} or a \textit{Project Manager}.
    \item \textbf{\textit{syscon}}: represents any system concepts such as \textit{engineering unit}, \textit{product}, \textit{hardware}, \textit{software}, etc. They mostly represent physical concepts.
    \item \textbf{\textit{opcon}}: represents operational concepts such as \textit{decision analysis process}, \textit{technology maturity assessment}, \textit{system requirements review}, etc.
    \item \textbf{\textit{seterm}}: represents generic terms that are frequently used in SE text and those that do not fall under \textit{syscon} or \textit{opcon} such as \textit{project}, \textit{mission}, \textit{key performance parameter}, \textit{audit} etc.
    \item \textbf{\textit{event}}: represents event-like information in SE text such as \textit{Pre-Phase A}, \textit{Phase A}, \textit{Phase B}, etc.
    \item \textbf{\textit{org}}: represents an organization such as \textit{`NASA'}, \textit{`aerospace industry'}, etc.
    \item \textbf{\textit{art}}: represents names of artifacts or instruments such as \textit{`AS1300'}
    \item \textbf{\textit{cardinal}}: represents numerical values such as \textit{`1'}, \textit{`100'}, \textit{'one'} etc.
    \item \textbf{\textit{loc}}: represents location-like entities such as \textit{component facilities} or \textit{centralized facility}.
    \item \textbf{\textit{mea}}: represents measures, features, or behaviors such as \textit{cost}, \textit{risk}, or \textit{feasibility}.
    
\end{enumerate}

\subsection{Abbreviations} Abbreviations are used frequently in SE text. We automatically extract abbreviations using simple pattern-matching around parentheses. Given below is a sample regex that matches most abbreviations in the SE handbook. \begin{Verbatim}[fontsize=\footnotesize]
r"\([ ]*[A-Z][A-Za-z]*[ ]*\)"
\end{Verbatim}
An iterative regex matching procedure using this pattern over the preceding words will produce the full phrase of the abbreviation. \textit{`A process to determine a system’s technological maturity based on Technology Readiness Levels (TRLs)'} produces the abbreviation \textbf{\textit{TRL}} which stands for \textbf{\textit{Technology Readiness Levels}}. \textit{`Define one or more initial Concept of Operations (ConOps) scenarios'} produces the abbreviation \textbf{\textit{ConOps}} which stands for \textbf{\textit{Concept of Operations}}. We pre-label these abbreviations as concept entities. Many of these abbreviations are also provided in the Appendix section of the handbook which is also extracted and used as concepts.

\subsection{Common-Knowledge Definitions}
Various locations of the handbook and the glossary provide definitions of several SE concepts. We collect these and compile a comprehensive definitions document which is also used for the concept recognition task. An example definition and its description is shown below:\\
\indent \textit{\textbf{Definition}: Acceptable Risk} \\ 
\indent \textit{\textbf{Description}: The risk that is understood and agreed to by the program/project, governing authority, mission directorate, and other customer(s) such that no further specific mitigating action is required.}

\subsection{CR Dataset Construction and Pre-processing}

Using python tools such as \textit{PyPDF2}, \textit{NLTK}, and \textit{RegEx} we build a pipeline to convert PDF to raw text along with extensive pre-processing which includes joining sentences that are split, removing URLs, shortening duplicate non-alpha characters, and replacing full forms of abbreviations with their shortened forms. We assume that the SE text is free of spelling errors. For the CR dataset, we select coherent paragraphs and full sentences by avoiding headers and short blurbs. Using domain keywords and a domain expert, we annotate roughly $3700$ sentences at the word-token level. An example is shown in Figure 2 and the unique tag count is shown in Table 1.

\begin{figure}
  \centering
    \includegraphics[width=8.1cm]{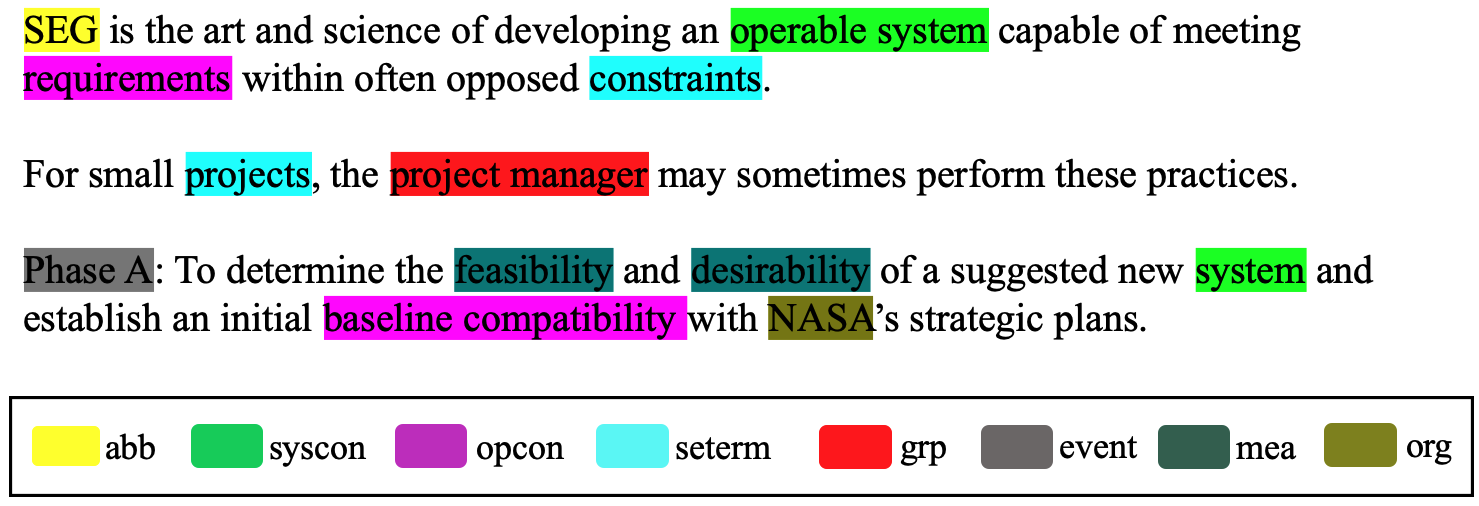}
 \caption{A Snippet of the concept-labelled dataset}
\end{figure}

\begin{table}
\footnotesize
\begin{center}
\begin{tabular}{|c|c||c|c||c|c| }
    \hline
O & 73944 & B-cardinal & 414 & I-grp & 132 \\
B-opcon & 5530 & B-abb & 354 & B-org & 87 \\
B-syscon & 1640 & B-event & 350 & I-seterm & 26 \\
B-seterm & 1431 & I-event & 218 & B-art & 17 \\
I-opcon & 1334 & I-syscon & 201 & I-org & 12 \\
B-mea & 1117 & I-abb & 156 & I-loc & 3 \\
B-grp & 499 & I-mea & 145 & B-loc & 2\\
\hline
\end{tabular}
\end{center}
 \caption{Unique Tag Count from the CR dataset} \label{table:unique_tag}
\end{table}

\subsection{Fine tuning with BERT}

Any language model can be used for the purpose of customizing an NER problem to CR. We choose to go with BERT \cite{bert} because of its general-purpose nature and usage of contextualized word embeddings. 

In the hand-labelled dataset, each word gets a label. The idea is to perform multi-class classification using BERT's pre-trained cased language model. We use pytorch transformers and hugging face as per the tutorial by \cite{bert-tut} which uses $BertForTokenClassification$. The text is embedded as tokens and masks with a maximum token length. This embedded tokens are provided as the input to the pre-trained BERT model for a full fine-tuning. The model gives an F1-score of $0.89$ for the concept recognition task. An 80-20 data split is used for training and evaluation. Detailed performance of the CR is shown in Table 2 and 3. Additionally, we also implemented CR using spaCy \cite{spacy} which also produced similar results.

\begin{table}[!htbp]
\footnotesize
\centering
\begin{tabular}{l|l|l|l|r|}
\cline{2-5}
& \textbf{precision} & \textbf{recall} & \textbf{f1-score} & \textbf{support} \\ \hline
\multicolumn{1}{|l|}{syscon} &  0.94   &  0.89   & 0.91 &  320                    \\ \hline
\multicolumn{1}{|l|}{opcon} &  0.87   &  0.91   & 0.89 &  1154                     \\ \hline
\multicolumn{1}{|l|}{seterm} &  0.98   &  0.94   & 0.96 &  287                     \\ \hline
\multicolumn{1}{|l|}{mea} & 0.91    &  0.90   & 0.90 &  248                     \\ \hline
\multicolumn{1}{|l|}{grp} &  0.94   &  0.93   & 0.94 &    89                   \\ \hline
\multicolumn{1}{|l|}{org} & 1.00    &   0.11  & 0.21 &  26                     \\ \hline
\multicolumn{1}{|l|}{cardinal} & 0.90    &  0.92   & 0.91 &  71                     \\ \hline
\multicolumn{1}{|l|}{event} &  0.71   & 0.78    & 0.76 &  77                    \\ \hline
\multicolumn{1}{|l|}{abb} &  0.82   &  0.58   & 0.68 &  79                     \\ \hline
\multicolumn{1}{|l|}{art} &  0.00   &   0.00  & 0.00 &  4                     \\ \hline
\multicolumn{1}{|l|}{loc} &  0.00   &   0.00  & 0.00 &  1                     \\ \hline
\multicolumn{1}{|l|}{\textbf{micro/macro-avg}} & \textbf{0.90}    & \textbf{ 0.88}   & \textbf{0.88} &  \textbf{2356}                     \\ \hline
\end{tabular}
\caption{Performance of different labels}
\end{table}

\begin{table}[!htbp]
\footnotesize
\begin{center}
 \begin{tabular}{||p{1.5cm}| p{1.5cm} | p{3.5cm}||} 
 \hline
 \textbf{F1-Score} & \textbf{Accuracy} &  \textbf{Accuracy without `O'-tag}\\ [0.5ex] 
 \hline\hline
 0.89 & 0.97 & 0.86  \\ 
 \hline
 \end{tabular}
  \caption{Overall Performance of CR; For fairness, we also provide the accuracy when the most common `O'-tag is excluded from the analysis.}
\end{center}
\end{table}

\section{RELATION EXTRACTION}

In this work, for relation extraction, we focus on hyponyms and verb phrase chunking. Hyponyms are more specific concepts such as \textit{earth} to \textit{planet} or \textit{rose} to \textit{flower}. Verb phrase chunking connects the named entities recognized by the CR model through verbs.

\subsection{Hyponyms from Definitions}
The definition document consists of $241$ SE definitions and their descriptions. We iteratively construct entities in increasing order of number of words in the definitions with the help of their parts-of-speech tags. This helps in creating \textit{subset-of} relation between a lower-word entity and a higher-word entity. Each root entity is lemmatized such that entities like \textit{processes} and \textit{process} appear only once. \\

\includegraphics[width=5.8cm]{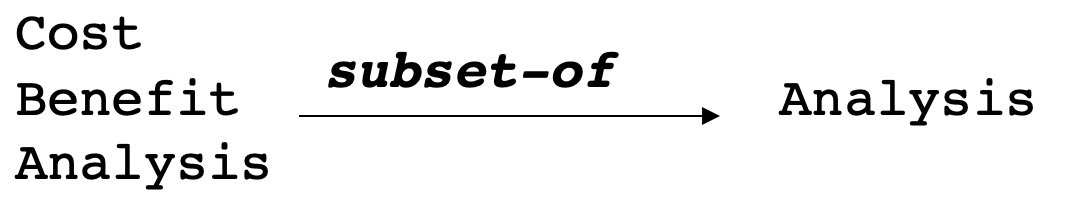}

\subsection{Hyponyms from POS tags}
Using the words (especially nouns) that surround an already identified named entity, more specific entities can be identified. This is performed on a few selected entity tags such as \textit{opcon} and \textit{syscon}. For example, consider the sentence \textit{`SE functions should be performed'}. \textit{`SE'} has tag \textit{NNP} and \textit{`functions'} has tag \textit{NNS}. We create a relation called \textbf{\textit{subset-of}} between \textit{`SE functions'} and \textit{`SE'}. \\

\includegraphics[width=6cm]{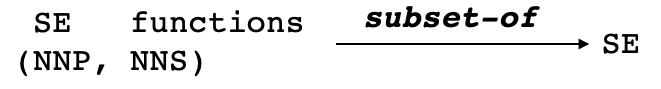}

\subsection{Relations from Abbreviations}
\begin{figure}[!htbp]
  \centering
    \includegraphics[width=7.8cm]{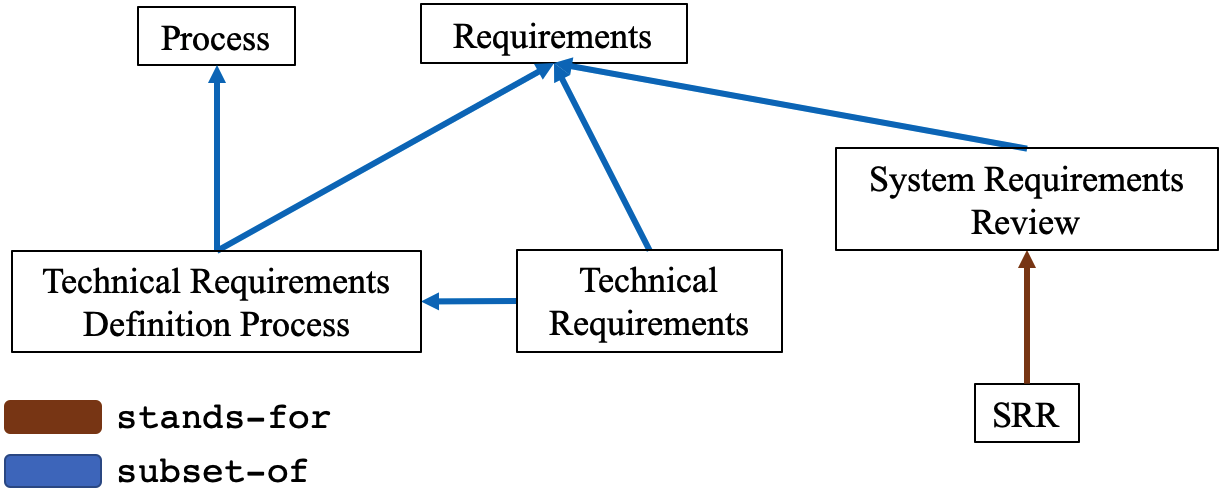}
 \caption{A snippet of the knowledge graph generated}\label{fig:kg}
\end{figure}

\includegraphics[width=5.7cm]{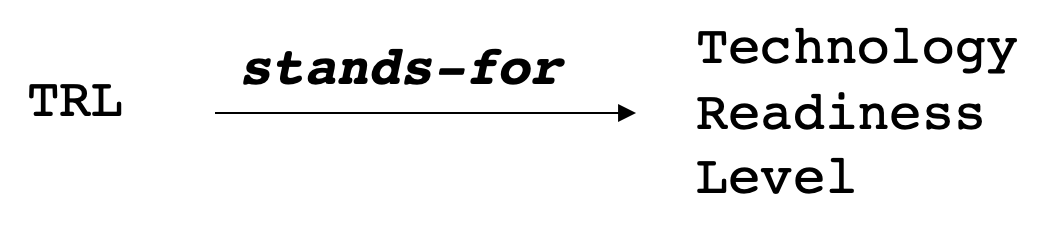}\\
Relations from abbreviations are simple direct connections between the abbreviation and its full form described in the abbreviations dataset. Figure \ref{fig:kg} shows a snippet of knowledge graph constructed using \textit{stands-for} and \textit{subset-of} relationships. Larger graphs are shown in the demo. \\

\subsection{Relation Extraction using Verb Phrase Chunking}

\begin{figure}[!htbp]
  \centering
    \includegraphics[width=8.3cm]{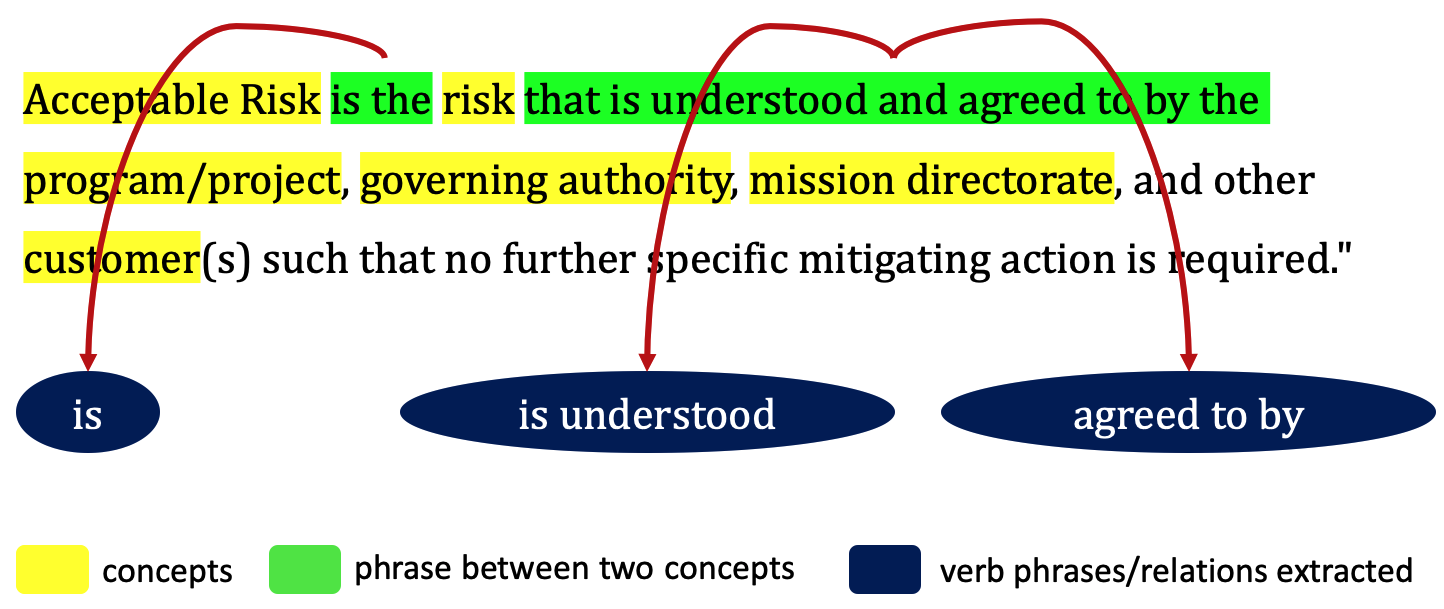}
 \caption{Relation Extraction using Verb Phrase}\label{fig:verb_phrase}
\end{figure}

Finally, we explore creating contextual triples from sentences using all the entities extracted using the CR model and entities from definitions. Only those phrases that connect two entities are selected for verb phrase extraction. Using NLTK's regex parser and chunker, a grammar such as 
\begin{Verbatim}[fontsize=\footnotesize]
VP: {(<MD>|<R.*>|<I.*>|<VB.*>|<JJ.*>|
<TO>)*<VB.*>+(<MD>|<R.*>|<I.*>|<VB.*>|
<JJ.*>|<TO>)*} 
\end{Verbatim} 
with at least one verb, can extract relation-like phrases from the phrase that links two concepts. An example is shown in Figure \ref{fig:verb_phrase}. Further investigation of relation extraction from SE handbook is left as future work.

\section{CONCLUSION AND FUTURE WORK}

We presented a common-knowledge concept extractor for the Systems Engineer's Virtual Assistant (SEVA) system and showed how it can be beneficial for downstream tasks such as relation extraction and knowledge graph construction. We construct a word-level annotated dataset with the help of a domain expert by carefully defining a labelling scheme to train a sequence labelling task to recognize SE concepts. Further, we also construct some essential datasets from the SE domain which can be used for future research. Future directions include constructing a comprehensive \textit{common-knowledge} relation extractor from SE handbook and incorporating such human knowledge into a more comprehensive machine-processable \textit{commonsense} knowledge base for the SE domain. 

\bibliographystyle{aaai}
\bibliography{references} 

\end{document}